\documentclass[11pt]{article} 
\usepackage{rldmsubmit,palatino}
\usepackage{graphicx}
\usepackage{subcaption} 
\usepackage{mymacros}
\usepackage{natbib}
\title{Scalable Multi-Agent Offline Reinforcement Learning and \\ the Role of Information}

\newtheorem{theorem}{Theorem}[section]

\newtheorem{lemma}[theorem]{Lemma}

\author{
Riccardo Zamboni\\ 
Politecnico di Milano\\
Milano, Italy \\
\texttt{riccardo.zamboni@polimi.it} \\
\And
Enrico Brunetti \\
Politecnico di Milano\\
Milano, Italy \\
\texttt{enrico.brunetti@polimi.it} \\
\AND
Marcello Restelli \\
Politecnico di Milano\\
Milano, Italy \\
\texttt{marcello.restelli@polimi.it}
}

\begin{document}

\maketitle

\begin{abstract}
Offline Reinforcement Learning (RL) focuses on learning policies solely from a batch of previously collected data. offering the potential to leverage such datasets effectively without the need for costly or risky active exploration. While recent advances in Offline Multi-Agent RL (MARL) have shown promise, most existing methods either rely on large datasets jointly collected by all agents or agent-specific datasets collected independently. The former approach ensures strong performance but raises scalability concerns, while the latter emphasizes scalability at the expense of performance guarantees. 
In this work, we propose a novel scalable routine for both dataset collection and offline learning. Agents first collect diverse datasets coherently with a pre-specified information-sharing network and subsequently learn coherent localized policies without requiring either full observability or falling back to complete decentralization. We theoretically demonstrate that this structured approach allows a multi-agent extension of the seminal Fitted Q-Iteration (FQI) algorithm~\citep{riedmilled2005fittedqiteration} to globally converge, in high probability, to $\epsilon$-optimal policies. The convergence is subject to error terms that depend on the informativeness of the shared information. Furthermore, we show how this approach allows to bound the inherent error of the supervised-learning phase of FQI with the mutual information between shared and unshared information. 
Our algorithm, SCAlable Multi-agent FQI (SCAM-FQI), is then evaluated on a distributed decision-making problem. The empirical results align with our theoretical findings, supporting the effectiveness of SCAM-FQI in achieving a balance between scalability and policy performance.
\end{abstract}



\startmain 

\section{Introduction}

Offline Reinforcement Learning (RL) was shown to be a promising way towards deploying RL in real-world scenarios where interaction with the environment is prohibitive, costly, or risky \cite{mopo}. 
For the many practical scenarios that involve multiple agents, recent results have shown that online RL algorithms can be scaled to multi-agent scenarios 
in a relatively straightforward way \cite{thesurprisingeffectiveness}. Surprisingly, little is known about how single-agent offline algorithms behave in the presence of multiple agents, and even less is known about which theoretical guarantees do transfer. Existing works either assume full observability, i.e., each agent accesses the states of the whole team, or partial observability, i.e., each agent accesses her own states only. While the former setting enforces good guarantees at the cost of several practical limitations \cite{bardemodelbased}, the latter one fails to provide any relevant performance guarantee \cite{Pan2021PlanBA}. To address these issues, we first propose a novel and scalable offline routine in which agents collect distinct datasets, yet they partially share their respective states according to a communication graph. We then propose a multi-agent adaptation of Fitted Q-Iteration (FQI) algorithm \citep{riedmilled2005fittedqiteration}, namely SCAlable Multi-agent FQI (\textbf{SCAM-FQI}), the first offline MARL algorithmic instantiation that is scalable both in training and in deployment, as both value functions and policies are supported or conditioned over the states of a restricted number of agents according to the communication graph. We describe how the proposed algorithm enjoys nice convergence guarantees that either explicitly or implicitly depend on the quality of the information shared between the agents. Finally, we couple these statements with some empirical corroboration over an (almost) real-case decision-making scenario consisting of an industrial production scheduling problem.

\vspace{-5pt}
\textbf{Notation.}~~In the following, we denote a set with a calligraphic letter $\mathcal A$ and its size as $|\mathcal A|$. We employ $\times_{i \in \mathcal \Ns} \mathcal A_i$ the Cartesian product over a finite set $\Ns$. For a (finite) set $\mathcal A = \{1,2,\dots, i, \dots, |\mathcal A|\}$, we denote with $-i = \Acal / \{i\}$ the set of all its elements out of the $i$-th one. The simplex on $\mathcal A$ is denoted as $\Delta(\mathcal A) := \{ p \in [0, 1]^{|\mathcal A|} | \sum_{a \in \mathcal A} p(a) = 1 \}$. In general, for $f : \mathcal A \rightarrow \mathbb R$ we denote the weighted p-norm as $\| f \|_{p,\nu}^p :=\E_{a \sim \nu}|f^p(a)| $ for a (probability) measure $\nu \in \Delta(\mathcal A)$, we denote the supremum-norm with$\| f \|_\infty$. Finally, we denote with $p \circ q$ the composition of two distributions.

\section{Problem Setting}
\label{problem_formulation}

In this work, we focus on \textbf{Markov Games}~\citep[MGs,][]{Littman1994}, which consist of a tuple $\mathcal M = \{\Ns, \mathcal S, \mathcal A, \P, r, \gamma, \mu\}$ with $\Ns$ being the set of agents interacting in the game, $\mathcal S$ and $\mathcal A$ be the joint state and action spaces respectively. $\P: \mathcal S \times \mathcal A \rightarrow \Delta(\mathcal S)$ is the (joint) transition model governing the game dynamics, while $r$ is the reward function assumed to be sum-decomposable over the agents, namely $r: \mathcal S \times \mathcal A \rightarrow [0, R^{\text{max}}], r(s,a) := \sum_{i \in \Ns} r_i(s,a)$. The state and action spaces are assumed to be factored over agents, namely $\mathcal S = \times_{i \in \Ns} \mathcal S_i$ and $\mathcal A = \times_{i \in \Ns} \mathcal A_i$. Finally, $\gamma$ and $\mu$ are the discount factor employed in the evaluation metric and the initial state distribution, respectively. We will consider the existence of an information-sharing graph $\mathcal G= (\Ns, \mathcal E)$. The vertex set $\Ns = \{1,2,...,|\Ns|\}$ denotes the set of agents and the edge set $\mathcal E$ prescribes the communication links among agents: we define $\Ns_i = \{j \in \Ns|(i,j) \in \mathcal E\}$ as the set of agents that share their state-space information with agent $i$. Thus, we denote the set of states of agents in $\Ns_i$ with $\mathcal S_{\Ns_i}$ and the global policy is $\pi(a|s)= \prod_{i \in \Ns}\pi_i(a_i|s_{\Ns_i})$, i.e., given global state s, each agent acts independently according to its own policy $\pi_i: \mathcal S_{\Ns_i} \rightarrow \Delta(\mathcal A_i)$. Additionally, the graph $\mathcal G$ induces for each agent the so-called \emph{local transitions and rewards}, namely 
$\P^i(s_{ \Ns_i}'|s_{ \Ns_i},a_i):  \mathcal S_{ \Ns_i}\times \mathcal A_i \rightarrow  \Delta( S_{ \Ns_i})$ and $ \bar{r}_{i}(s_{ \Ns_i}, a_i) : \mathcal S_{\Ns_i}\times \mathcal A_i \rightarrow  \mathbb R $. Finally, joint states are distributed according to $d^\pi_\mu \in \Delta_{\Ss}$, where $d^\pi (s) = (1-\gamma)\sum_{t =0}^\infty \gamma^t Pr (s_t = s|\pi,\mu)$. As a result, we can define the \emph{decentralized Q-functions} as:
\vspace{-5pt}
\begin{equation}
     q_i^{\pi_i}(s_{\Ns_i},a_i) := \E_{\pi^i, \P^i}[\sum_t^\infty \gamma^t \bar r_{i}(s_{{ \Ns_i}, t},a_{i,t})|(s_{{ \Ns_i}, 0},a_{i,0})=(s_{\Ns_i},a_i)], \quad  q^\pi (s,a) := \sum_{i \in \Ns}  q_i^{\pi_i}(s_{\Ns_i},a_i)
     \vspace{-5pt}
\end{equation}
These will be compared against their \emph{centralized} counterparts:
\vspace{-8pt}
\begin{equation}
    Q_i^{\pi}(s_{ \Ns_i},a_i) := \E_{\pi, \P}[\sum_t^\infty \gamma^t\bar r_{i}(s_{{ \Ns_i}, t},a_{i,t})|(s_{{ \Ns_i}, 0},a_{i,0})=(s_{\Ns_i},a_i)], \quad Q^\pi (s,a) := \sum_{i \in \Ns} Q_i^{\pi}(s_{ \Ns_i},a_i)
    \vspace{-5pt}
\end{equation}
Finally, we will introduce the \emph{optimal Bellman Operators} for decentralized functions $f_i :\mathcal S_{\Ns_i} \times \mathcal A_i \rightarrow \mathbb R$ and centralized ones $f: \mathcal S \times \mathcal A \rightarrow \mathbb R$ respectively as:
\begin{align}
    &\bar{\mathcal T}^if_i(s_{ \Ns_i},a_i) = \bar r_i(s_{ \Ns_i}, a_i) + \gamma  \E_{s_{ \Ns_i}' \sim \P^i(\cdot|s_{ \Ns_i}, a_i)}\max_{a_i'}f_i(s_{ \Ns_i}',a_i'),
    &\mathcal T f(s,a) = r(s,a) + \gamma \E_{s' \sim \P(\cdot|s,a)} \max_{a'}f(s',a') \label{eq:bellmancentralized}
\end{align}\vspace{-14pt}

where $\bar{\mathcal  T} f(s,a) := \sum_{i \in \Ns}  \bar {\mathcal T}^if_i(s_{ \Ns_i},a_i)$. The objective will be defined in centralized terms, that is as learning the \emph{optimal policy} $\pi^\star(\cdot|s)\in \argmax_\pi \E_{a \sim\pi} Q^\star (s,a) $ with $Q^\star$ being the fixed point of the centralized optimal Bellman Operator.

\section{Towards Scalable Offline MARL}
\label{performances}

In the following, we propose an offline learning approach that integrates a decentralized architecture with a structured communication network. Each agent collects a distinct dataset comprising its own local information \emph{and} limited (but potentially relevant) information from a subset of other agents, and subsequently learns a policy conditioned \emph{only} on this subset. Intuitively, restricting the information accessible to each agent introduces a potential performance trade-off. We will show that the degradation in performance is proportional to the extent to which the local transitions and reward approximate the true underlying dynamics. Crucially, omitting information that is either weakly relevant or highly redundant, quantified via Conditional Mutual Information (CMI), does not significantly impact the overall error. This selective conditioning also enables the exploitation of substantially larger per-agent datasets, thus improving sample efficiency without compromising performance guarantees.
\paragraph{Data Collection:}~~Each agent will have access to different data-set $\mathcal D_i = \{(s_{\Ns_i,t},a_{i,t},\bar r_{i,t},s'_{\Ns_i,t})\}$; in other words, agents will only have to collect their own experience and share it according to $\mathcal G$, without any joint collection happening. For the sake of the theoretical analysis, we will then assume that there exists a joint dataset $\mathcal D = \{(s_t,a_t,r_t,s'_t)\}$ collected according to $(s,a) \sim \nu \in \Delta(\mathcal S \times \mathcal A)$, $r = r(s,a)$, $s' \sim \P(\cdot|s,a)$ and that it enjoys a \emph{Concentrability Assumption}: $
    \forall \pi, \forall (s,a), \;\exists\; C:\; \frac{d^\pi(s,a) }{ \nu(s,a)} \leq C 
$. Crucially, the actual size of the  datasets will depend on $\mathcal G$, and that, in general, $|\mathcal D| << |\mathcal D_i|$.
\paragraph{Offline Iteration:}~~The proposed algorithm SCAM-FQI consists of a scalable multi-agent instantiation of FQI~\citep{riedmilled2005fittedqiteration}; it performs the following simple iteration: starting with some function $\hat q^0_i \in \mathcal F_i$ over $\mathcal F_i = \{f : \mathcal S_{\Ns_i} \times \mathcal A_i \rightarrow [0, V^{\text{max}}]\},  V^{\text{max}} = \frac{R^{\text{max}}}{1-\gamma}$, each agent at iteration $k$ independently performs the following updates:
\vspace{-5pt}
\begin{equation} \label{fqi_iteration}
    \hat q_i^{k} \in  \argmin_{f \in \mathcal F_i} \frac{1}{|\mathcal D_i|}\sum_{\mathcal D_i} \Big(f(s_{ \Ns_i, t}, a_{i,t}) -\bar{r}_{i,t} -\gamma \max_{a_i'}\hat q^{k-1}_i (s'_{ \Ns_i, t}, a_{i}')\Big)^2, \quad \pi_i^k(a_i|s_{ \Ns_i}) := \argmax_{a_i} \hat q^k_i(s_{ \Ns_i}, a_i)
\end{equation}
\vspace{-10pt}

\textbf{Convergence Guarantees:}~~Now, we show with a proof following standard techniques~\citep{Agarwal2019ReinforcementLT} that SCAM-FQI leverages the information-sharing graph so as to maintain good performance guarantees. First of all:
\begin{lemma}\label{lem:lem1}
Define the value functions $V^\pi_\mu:= \E_{s \sim d^\pi_\mu, a \sim \pi(\cdot|s)}Q^\pi(s,a)$, assume that for each iteration $k\geq 0$, it holds $\forall i \in \Ns$ that $\norm{ \hat q^{\pi^k_{i}} - \bar{\mathcal T}^i \hat q^{\pi_i^{k-1}}}_{2,\nu} \leq \epsilon$ then for any initial distribution $\mu \in \Delta(\mathcal S)$ it holds that  
\vspace{-6pt}
\begin{equation}
    V^\star_\mu - V^{\pi^k}_\mu \leq\frac{ 2\gamma^{k-1}}{1-\gamma} (\gamma V^\text{max} + \sqrt{C}\epsilon_r + \frac{ \sqrt{C}\gamma}{1-\gamma}\epsilon_\P) + \frac{2N\sqrt{C}}{(1-\gamma)^2}\epsilon,
\end{equation}
with $\epsilon_r := \sum_{i \in \Ns}\norm{r_i - \bar r_i}_{1,\nu}, \epsilon_\P := \sum_{i \in \Ns} \norm{\P - \P^i}_{1,\nu}$.
\end{lemma}
\proof{
We start with the Performance Difference Lemma between joint functions against the optimal policy $\pi^\star$\footnote{Here we simplified the notation by denoting $A^\star(s, \pi^k(s)) := \E_{a \sim \pi^k(\cdot|s)}A^\star(s,a)$ and dropping the dependency over the fixed initial distribution $\mu$.}:
\begin{align*}
    (1-\gamma)(V^\star_\mu - V^{\pi^k}_\mu) &= \E_{s \sim d^{\pi^k}}[Q^\star(s, \pi^\star(s))-Q^\star(s, \pi^k(s))] 
    \leq\E_{s \sim d^{\pi^k}}[Q^\star(s, \pi^\star(s))-\hat q^{\pi^k}(s,\pi^\star(s)) + \hat q^{\pi^k}(s,\pi^k(s)) - Q^\star(s, \pi^k(s))]  \\
    &\leq \norm{Q^\star - \hat  q^{\pi^k} }_{1,d^{\pi^k}\circ \pi^\star} + \norm{Q^\star - \hat q^{\pi^k} }_{1,d^{\pi^k}\circ \pi^k}\\
\intertext{Now, for a generic term we have that $\norm{Q^\star - \hat q^{\pi^k} }_{1,\beta} \leq \overset{(a)}{\norm{Q^\star - {\mathcal T} \hat q^{\pi^{k-1}}}_{1,\beta}}+\overset{(b)}{\norm{ \hat  q^{\pi^k} - \bar{\mathcal T} \hat q^{\pi^{k-1}}}_{1,\beta}}+\overset{(c)}{\norm{\mathcal T\hat q^{\pi^{k-1}}- \bar{\mathcal T}\hat q^{\pi^{k-1}}}_{1,\beta}}$, and }
\norm{Q^\star - {\mathcal T} \hat q^{\pi^{k-1}}}_{1,\beta}&\overset{(a)}{\leq}  \gamma \E_{(s,a)\sim \beta}\big|\E_{s'\sim \P(\cdot|s,a)}\max_{a'}Q^\star(s',a')- \max_{a'} \hat  q^{\pi^{k-1}}(s',a')\big|\\
&\leq \gamma \norm{Q^\star -\hat q^{\pi^{k-1}} }_{1,\beta''}, \beta''(s',a') = \sum_{s,a}\beta(s,a)\P(s'|s,a)\mathbf{1}\{a' = \argmax_a|Q^\star(s',a)-\hat q^{\pi^{k-1}}(s',a)|\}\\
\norm{\mathcal T\hat q^{\pi^{k-1}}- \bar{\mathcal T}\hat q^{\pi^{k-1}}}_{1,\beta} &\overset{(b)}{=} \E_{(s,a)\sim \beta}\big|\sum_i  r_i(s,a)+\gamma \E_{s'\sim \P(\cdot|s,a)} \max_{a'} \hat q^{\pi^{k-1}}(s',a')
- \sum_i \bar r_i(s_{\Ns_i},a_i)-\gamma\E_{s'_{ \Ns_i}\sim  \P^i(\cdot|s_{ \Ns_i},a_i)}\max_{a_i'}\hat q^{\pi^{k-1}_i}_i(s'_{\Ns_i},a_i'))\big|\\
&\leq \sum_i\norm{r_i - \bar r_i}_{1,\beta} + \frac{\gamma}{1-\gamma}\norm{\P -  \P^i}_{1,\beta} \\
\norm{ \hat  q^{\pi^k} - \bar{\mathcal T} \hat q^{\pi^{k-1}}}_{1,\beta} &\overset{(c)}{\leq}\norm{ \hat  q^{\pi^k} - \bar{\mathcal T} \hat q^{\pi^{k-1}}}_{2,\beta} \leq \sum_i \sqrt{C}\norm{ \hat q^{\pi_i^{k}} - \bar{\mathcal T}^i \hat q^{\pi_i^{k-1}}}_{2,\nu}\\
\intertext{It follows that by telescoping up to $k=0$ after defining $\epsilon_r := \sum_i\norm{r_i - \bar r_i}_{1,\nu}, \epsilon_\P := \sum_i \norm{\P - \P^i}_{1,\nu}$, we get:}
(1-\gamma)(V^\star_\mu - V^{\pi^k}_\mu) &\leq 2\gamma\norm{Q^\star -\hat q^{\pi^{k-1}} }_{1,\beta''} + \sum_i 2 \sqrt{C} \norm{r_i - \bar r_i}_{1,\nu} + \frac{2 \sqrt{C} \gamma}{1-\gamma}\norm{\P - \P^i}_{1,\nu}+ 2\sqrt{C}\norm{ \hat q^{\pi_i^{k}} - \bar{\mathcal T}^i \hat q^{\pi_i^{k-1}}}_{2,\nu}\\ 
&\leq2\gamma^{k-1}(\gamma V^\text{max} + \sqrt{C}\epsilon_r + \frac{ \sqrt{C}\gamma}{1-\gamma}\epsilon_\P) + 2N\sqrt{C}\sum_{t=0}^{k-1} \gamma^t\epsilon \leq 2\gamma^{k-1} (\gamma V^\text{max} + \sqrt{C}\epsilon_r + \frac{ \sqrt{C}\gamma}{1-\gamma}\epsilon_\P) + \frac{2N\sqrt{C}}{1-\gamma}\epsilon
\end{align*}
}
Now, even assuming simple least-squares estimation, generalization bounds hold for the $\epsilon$ error term of Lem.~\ref{lem:lem1}:

\begin{lemma}[Least Squares Generalization Bound \cite{kakaderandomdesign}]\label{lemma:generalization}
    With probability at least $1 - \delta$ for all $k = 0 \dots K$ and $i \in \Ns$, it holds for the solution of the LS problem:
    \vspace{-5pt}
\begin{equation}
    \norm{ \hat q^{\pi_i^{k}} - \bar{\mathcal T}^i \hat q^{\pi_i^{k-1}}}_{2,\nu}^2 \leq \frac{22(V^\text{max})^2 \ln(|\mathcal F^i|KN/\delta)}{|D_i|} + 20 \epsilon_{\text{INH}},
\end{equation}
where $\epsilon_{\text{INH}} = \min_{f' \in \mathcal F_i}\max_{f \in \mathcal F_i}\|f' - \bar{\mathcal T}^i f\|^2_{2,\nu}$ is referred to as inherent Bellman Error.
\end{lemma}
\vspace{-5pt}
By combining all the previous results together, we can finally state the following:

\begin{theorem}[SCAM-FQI Convergence Guarantees] Define the value functions $V^\pi_\mu:= \E_{s \sim d^\pi_\mu, a \sim \pi(\cdot|s)}Q^\pi(s,a)$, assume that $\forall i,j \in \Ns \mathcal F_i = \mathcal F_j = \mathcal F, D_i=D_j=D, i \neq j$, fix $K \in \mathbb N^+$. The SCAM-FQI Iteration of Eqs.~\eqref{fqi_iteration} guarantees that with probability $1 - \delta$ for any initial distribution $\mu \in \Delta(\mathcal S)$ it holds
\vspace{-5pt}
\begin{equation}
    V^\star_\mu - V^{\pi^K}_\mu \leq\frac{ 2\gamma^{K-1}}{1-\gamma} (\gamma V^\text{max} + \sqrt{C}\epsilon_r + \frac{ \sqrt{C}\gamma}{1-\gamma}\epsilon_\P) + \frac{2N}{(1-\gamma)^2}\Bigg(\sqrt{\frac{22C(V^\text{max})^2 \ln(|\mathcal F|KN/\delta)}{|D|} } + \sqrt{20\epsilon_{\text{INH}}}\Bigg).
\end{equation}
\end{theorem}
\vspace{-8pt}
In other words, we are able to almost fully recover the convergence guarantees attained in the single-agent cases out of some core differences: $(i)$ the two error terms $\epsilon_r$ and $\epsilon_\P$ represent the bias induced by the per-agent local transitions and reward functions, and they are indeed controlled by the ability of the network $\mathcal G$ to recover the true transition and reward models; $(ii)$ the second term apparently scales linearly with the number of agents, yet one should notice that the nature of the communication network indirectly induces a trade-off on it, as the size of the actual per-agent dataset $|\mathcal D_i|$ will increase as the set $\Ns_i$ decreases; $(iii)$ one more refined analysis of the last supervised-learning related term reveals a trade-off on the information-sharing network as well:
\vspace{-5pt}

\paragraph{The Role of Information in the Supervised-Learning Error.}~~The proof of Lemma~\ref{lemma:generalization} builds on the simplifying assumption that at every step $k$ it is possible to (union) bound the error induced by the optimal solution $\|\tilde q^k - \bar{\mathcal T}^i \hat q^{k-1}_i\|^2_{2,\nu} \leq \epsilon_{\text{INH}}$ with $\tilde q^k = \arginf_{f \in \mathcal F_i}\|f - \bar{\mathcal T}^i \hat q^{k-1}_i\|^2_{2,\nu}$ and $\bar{\mathcal T}^i \hat q^{k-1}_i$ being the Bellman optimum. Yet, it is again possible to leverage the structure induced by the information-sharing network $\mathcal G$ to recover the following: 
\begin{lemma}[Th.1 of~\cite{Beraha2019FeatureSV}]
    Consider a fixed functional space $\mathcal F_i$ supported on $\mathcal S_{\Ns_i}$ induced by $\mathcal G$ and a full representation $\mathcal F_i^{\star}$ supported on $\mathcal S$, then by defining the CMI between two representations $\mathbb I(\mathcal N_i) = \mathbb I(Y;s_{ \Ns_{-i}}|s_{ \Ns_{i}})$, with $Y$ being the target of the regression problem, and by defining the irreducible error under the full representation $\sigma^2 = \norm{ \tilde q^{k,\star} - \bar{\mathcal T}^i \hat q^{k-1}_i}_{2,\nu}^2$, it follows that $
        \|\tilde q^k - \bar{\mathcal T}^i \hat q^{k-1}_i\|^2_{2,\nu}  \leq \sigma^2 + \mathbb I(\mathcal N_i). 
    $
\end{lemma}
\vspace{-8pt}
This Lemma states that by employing reduced-size information, one can bound the resulting error by the CMI between the target and the unused states, conditioned on the used ones. In other words, whenever a subset of states has low relevancy or high redundancy with the ones employed, that is, a low CMI, such states can be safely removed from the information-sharing graph without significantly affecting the performance guarantees of the SCAM-FQI.

\section{Empirical Corroboration: Multi-Agent Production Scheduling}
\label{experiments}

The decision-making problem we corroborated SCAM-FQI on consists of a \emph{Multi-Agent Production Scheduling} Problem: a set of agents represents a production plant and aims to route products around so as to perform the productive operations required by each product, the last required operation consisting in removing the product from the plant. Each productive operation can be performed by just a subset of the agents, and each agent is connected only to a subset of others in a (dense) network. The \emph{state} of each agent corresponds to the list of products currently in it, and the remaining production operations for each of them; the \emph{actions} of each agent correspond to the link through which to move the products currently on the agent
; the objective is to minimize the overall production time, which means trying to produce all the required products as fast as possible; coherently, the \emph{reward} to each action was designed to be the negative time elapsed before an agent saw the product again, or the negative time until production if this did not happen again. As for the communication network $\mathcal G$, each agent was allowed to access the states of the agents whose distance on the plant was less than an integer $d$. Additionally, we considered the possibility of sharing not the \emph{full} states but a \emph{compressed} version, consisting of the mere information of whether the agent was currently busy on a product or not. The employed dataset contained $1k$ episodes of production collected through uniform stochastic policies. The regressors used in the supervised learning phase consisted of Extremely Randomized Trees Regressors, and the extracted policies consisted of $0.15$-greedy policies.

\begin{figure}[!]
    \centering
    \begin{subfigure}[b]{0.49\textwidth}
        \centering
        \includegraphics[width=\textwidth]{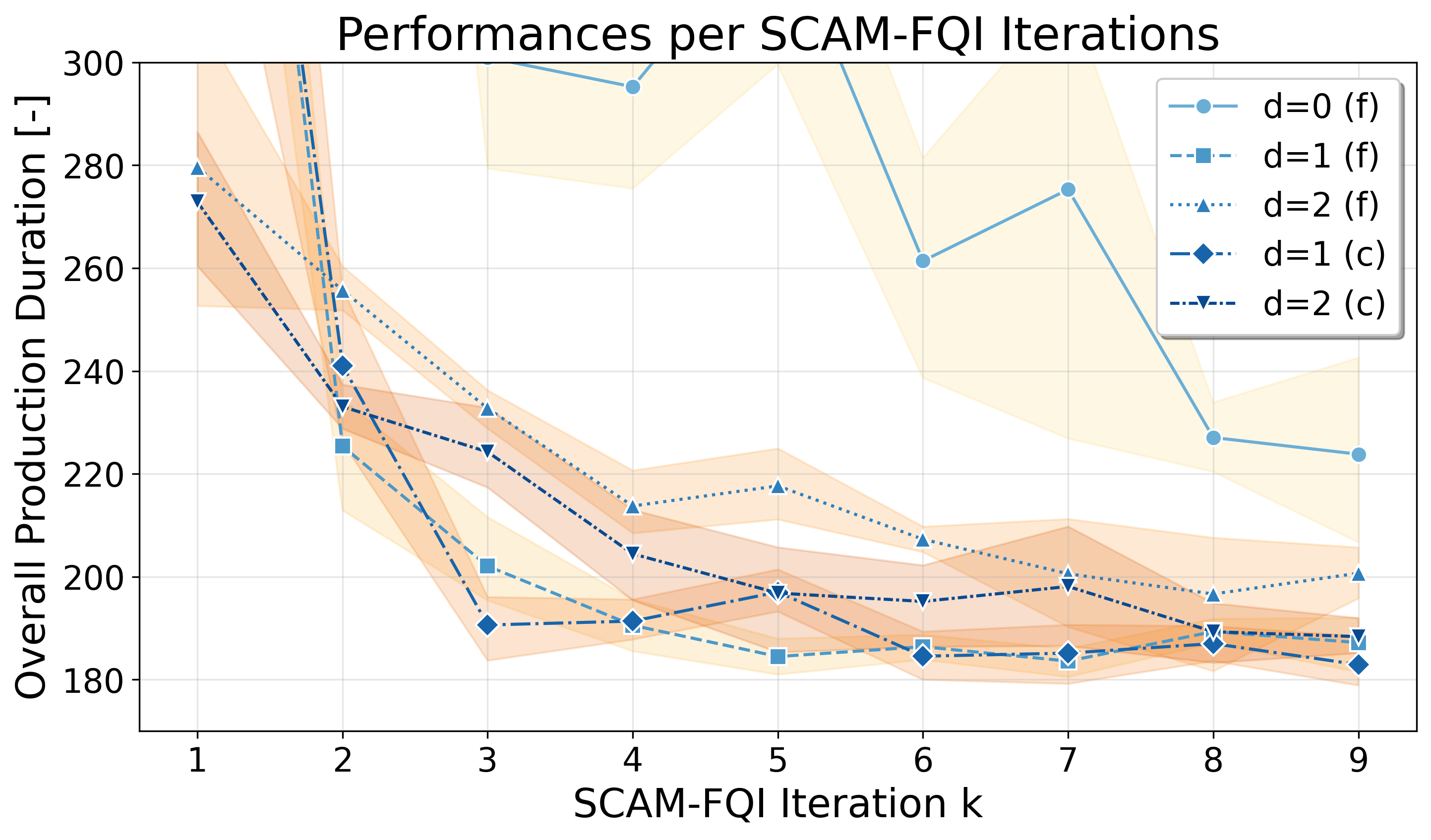}
        \caption{3 Products, each requiring different productive operations.}
        \label{fig:sub1}
    \end{subfigure}
    \hfill
    \begin{subfigure}[b]{0.49\textwidth}
        \centering
        \includegraphics[width=\textwidth]{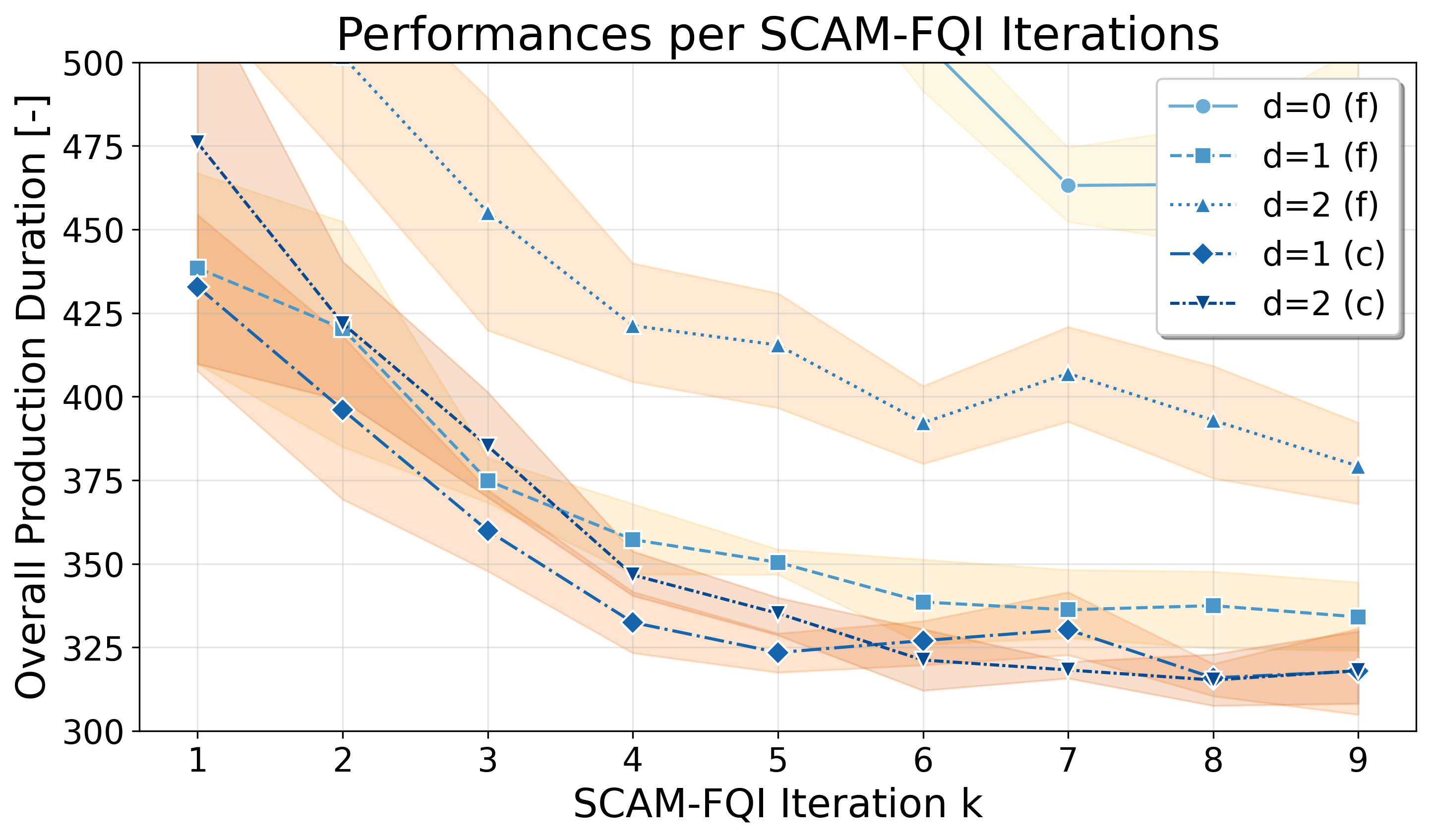}
        \caption{5 Products, each requiring the same productive operations.}
        \label{fig:sub2}
    \end{subfigure}
    \caption{$4\times5$ Densely Connected Agents Layout. The time (in ticks) required to perform all the required operations is plotted against SCAM-FQI iterations (\emph{lower is better}). The legend provides insights on the values of $d$ followed by whether the shared information was full (f) or compressed (c). Confidence intervals are bootstrapped $95\%$ CI over 5 seeds.}
    \label{fig:main}
    \vspace{-5pt}
\end{figure}
\vspace{-5pt}
\paragraph{Comments on the Empirical Corroboration.}~~As expected from the analysis in the previous Section, both the reported experiments show a relevant and common trend: the information-sharing among agents does induce a trade-off between the quality of the learned policies and the speed of convergence; a higher degree of sharing improves the performances while compressed forms of sharing allows for faster convergence while not harming the overall performances; however, over-sharing ends up causing a degradation of performances: the shared information increases the variance of the dataset without adding any value in terms of reduced bias, eventually affecting the learning negatively.
\vspace{-15pt}
\section{Conclusions}
\label{conclusions}

In this work, we first introduced a novel routine for multi-agent offline data collection, by building different per-agent datasets based on an information-sharing graph rather than centralized or decentralized datasets. We then proposed an algorithm, \textbf{SCAM-FQI}, that takes explicit advantage of such structure to learn scalable decentralized policies. Furthermore, we provided theoretical guarantees of SCAM-FQI convergence in terms of the overall quality of the information shared between the agents. Finally, we provided some empirical corroboration of the trade-off induced by the quality of the information sharing by testing SCAM-FQI against a distributed production scheduling problem. We believe that our work can be a crucial step in the direction of developing scalable solutions in a principled way to yet more practical settings. \vspace{-12pt}
\paragraph{Comments on Practical Applicability.}~~The proposed approach integrates a decentralized architecture with a structured communication network. This design mitigates the need for full-state observability at deployment, offering a clear advantage over fully centralized solutions. However, it presupposes the availability of an efficient communication protocol in fully decentralized environments. Consequently, the practical applicability of the method is best suited to high-dimensional, communication-efficient scenarios; settings where centralized approaches struggle to scale, but where a communication infrastructure can be feasibly deployed.

\vspace{-12pt}
\bibliographystyle{abbrv} 
\bibliography{biblio}

\begin{thebibliography}{1}

\bibitem{Agarwal2019ReinforcementLT}
A.~Agarwal, N.~Jiang, and S.~M. Kakade.
\newblock Reinforcement learning: Theory and algorithms.
\newblock In {\em Reinforcement Learning: Theory and Algorithms}, 2019.

\bibitem{bardemodelbased}
P.~Barde, J.~Foerster, D.~Nowrouzezahrai, and A.~Zhang.
\newblock A model-based solution to the offline multi-agent reinforcement learning coordination problem.
\newblock In {\em Proceedings of the 23rd International Conference on Autonomous Agents and Multiagent Systems}, AAMAS '24, page 141–150, 2024.

\bibitem{Beraha2019FeatureSV}
M.~Beraha, A.~M. Metelli, M.~Papini, A.~Tirinzoni, and M.~Restelli.
\newblock Feature selection via mutual information: New theoretical insights.
\newblock {\em 2019 International Joint Conference on Neural Networks (IJCNN)}, pages 1--9, 2019.

\bibitem{kakaderandomdesign}
D.~Hsu, S.~M. Kakade, and T.~Zhang.
\newblock Random design analysis of ridge regression.
\newblock {\em Found. Comput. Math.}, 14(3):569–600, June 2014.

\bibitem{Littman1994}
M.~L. Littman.
\newblock Markov games as a framework for multi-agent reinforcement learning.
\newblock In W.~W. Cohen and H.~Hirsh, editors, {\em Machine Learning Proceedings 1994}, pages 157--163. Morgan Kaufmann, San Francisco (CA), 1994.

\bibitem{Pan2021PlanBA}
L.~Pan, L.~Huang, T.~Ma, and H.~Xu.
\newblock Plan better amid conservatism: Offline multi-agent reinforcement learning with actor rectification.
\newblock {\em ArXiv}, abs/2111.11188, 2021.

\bibitem{riedmilled2005fittedqiteration}
M.~Riedmiller.
\newblock Neural {F}itted {Q} {I}teration.
\newblock In J.~Gama, R.~Camacho, P.~B. Brazdil, A.~M. Jorge, and L.~Torgo, editors, {\em Machine Learning: ECML 2005}, pages 317--328, Berlin, Heidelberg, 2005. Springer Berlin Heidelberg.

\bibitem{thesurprisingeffectiveness}
C.~Yu, A.~Velu, E.~Vinitsky, J.~Gao, Y.~Wang, A.~Bayen, and Y.~WU.
\newblock The surprising effectiveness of ppo in cooperative multi-agent games.
\newblock In {\em Advances in Neural Information Processing Systems}, volume~35, pages 24611--24624, 2022.

\bibitem{mopo}
T.~Yu, G.~Thomas, L.~Yu, S.~Ermon, J.~Y. Zou, S.~Levine, C.~Finn, and T.~Ma.
\newblock Mopo: Model-based offline policy optimization.
\newblock In H.~Larochelle, M.~Ranzato, R.~Hadsell, M.~Balcan, and H.~Lin, editors, {\em Advances in Neural Information Processing Systems}, volume~33, pages 14129--14142, 2020.

\end{thebibliography}

\end{document}